\documentclass{article}
\usepackage{ijcai19}

\usepackage[utf8]{inputenc}

\date{December 2018}

\usepackage{times}
\usepackage{soul}
\usepackage{url}
\usepackage[hidelinks]{hyperref}
\usepackage[utf8]{inputenc}
\usepackage[small]{caption}
\usepackage{graphicx}
\usepackage{algorithm}
\usepackage{algorithmic}
\usepackage{enumitem}
\usepackage{amsfonts}
\usepackage{dsfont}
\usepackage{amssymb}
\usepackage{amsmath}
\usepackage{amsfonts}
\usepackage{bm}
\usepackage{xspace}
\usepackage{helvet}
\usepackage{linguex}
\usepackage{xcolor}

\DeclareMathOperator*{\concat}{\scalebox{1}[1.5]{$\parallel$}}

\newcommand{\mname}{\texttt{G-BERT}\xspace}
\newcommand{\mnamengnp}{$\texttt{G-BERT}_{G^-,P^-}$\xspace}
\newcommand{\mnameng}{$\texttt{G-BERT}_{G^-}$\xspace}
\newcommand{\mnamenp}{$\texttt{G-BERT}_{P^-}$\xspace}

\title{Pre-training of Graph Augmented Transformers for Medication Recommendation}

\author{Junyuan Shang$^{1,3}$ \and Tengfei Ma$^2$ \and Cao Xiao$^1$ {\normalfont and} Jimeng Sun$^3$ \affiliations
$^1$Analytics Center of Excellence, IQVIA, Cambridge, MA, USA \\
$^2$IBM Research AI, Yorktown Heights, NY, USA\\
$^3$Georgia Institute of Technology, Atlanta, GA, USA 
\emails
junyuan.shang@iqvia.com, Tengfei.Ma1@ibm.com, cao.xiao@iqvia.com, jsun@cc.gatech.edu}

\begin{document}

\maketitle

\begin{abstract}
Medication recommendation is an important healthcare application.  It is commonly formulated as a temporal prediction task. Hence, most existing works only utilize longitudinal electronic health records (EHRs) from a small number of patients with multiple visits ignoring a large number of patients with a single visit ({\it selection bias}). Moreover, important {\it hierarchical knowledge} such as diagnosis hierarchy is not leveraged in the representation learning process. To address these challenges, we propose \mname,  a new model to combine the power of \underline{G}raph Neural Networks (GNNs) and \underline{BERT} (Bidirectional Encoder Representations from Transformers) for medical code representation and medication recommendation. We use  GNNs to represent the internal hierarchical structures of medical codes. Then we integrate the GNN representation into a transformer-based visit encoder and pre-train it on EHR data from patients only with a single visit. The pre-trained visit encoder and representation are then fine-tuned for downstream predictive tasks on longitudinal EHRs from patients with multiple visits. \mname is the first to bring the language model pre-training schema into the healthcare domain and it achieved state-of-the-art performance on the medication recommendation task.

\end{abstract}

\section{Introduction}
The availability of massive electronic health records (EHR) data and the advances of deep learning technologies have provided  unprecedented resource and opportunity for predictive healthcare, including the computational medication recommendation task. 
A number of deep learning models were proposed to assist doctors in making medication recommendation ~\cite{doi:10.1093/jamia/ocy068,shang2018gamenet,Baytas:2017:PSV:3097983.3097997,NIPS2018_7706,ma2018health}.  They often learn representations for medical entities (e.g., patients, diagnosis, medications) from patient EHR data, and then use the learned representations to predict  medications  that are suited to the patient's health condition.\\

\begin{figure}[t]
\centering
\includegraphics[width=0.85\linewidth]{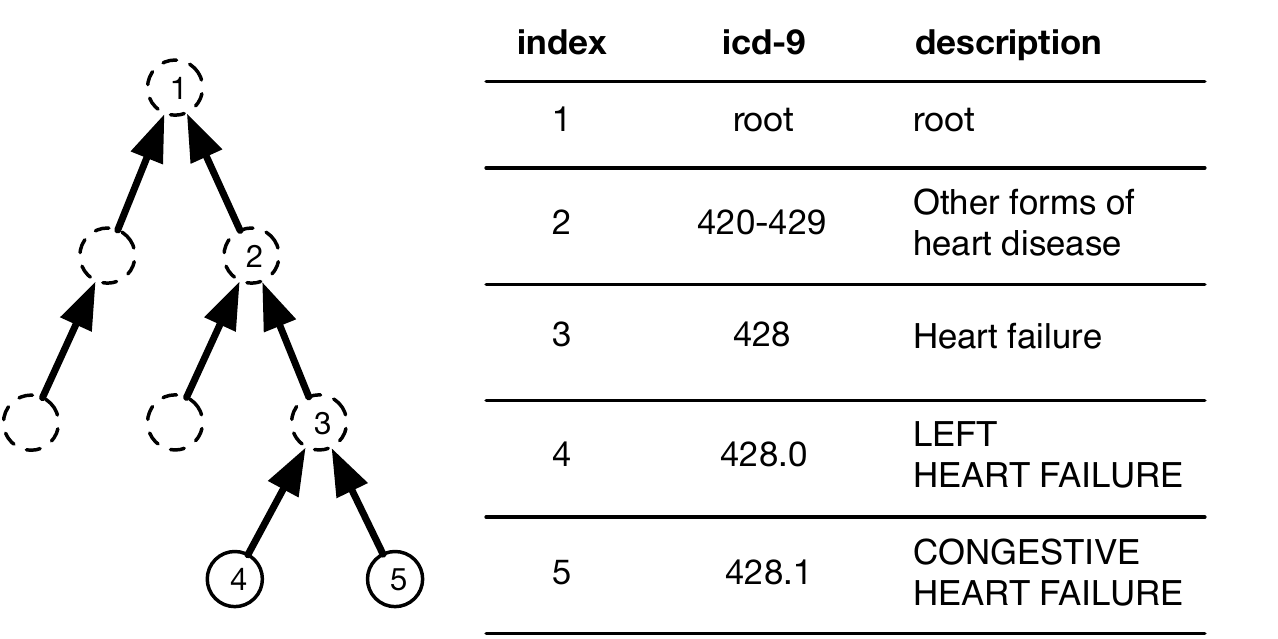}
\caption{Graphical illustration of ICD-9 Ontology.}
\label{fig:ontology}
\end{figure}

To provide effective medication recommendation, it is important to learn accurate representation of medical codes.
Despite that various considerations were handled in previous works for improving medical code representations~\cite{ma2018health,Baytas:2017:PSV:3097983.3097997,NIPS2018_7706}, there are two limitations with the existing work: 
\begin{enumerate}
    \item {\bf Selection bias:} Data that do not meet training data criteria are discarded before model training. For example, a large number of patients who only have one hospital visit were discarded from training in~\cite{shang2018gamenet}.
    \item  {\bf Lack of hierarchical knowledge:} For medical knowledge such as diagnosis code ontology (Figure.~\ref{fig:ontology}), their internal hierarchical structures were rarely embedded in their original graph form when incorporated into representation learning.
\end{enumerate}


To mitigate the aforementioned limitations, we propose \mname that combines the pre-training techniques and graph neural networks for better medical code representation and medication recommendation. \mname is enabled and demonstrated by the following technical contributions: 
\begin{enumerate}
\item \textit{Pre-training to leverage more data:} 
Pre-training techniques, such as ELMo~\cite{peters2018deep}, OpenAI GPT~\cite{radford2018improving} and BERT~\cite{devlin2018bert}, have demonstrated a notably good performance in various natural language processing tasks. These techniques generally train language models from unlabeled data, and then adapt the derived representations to different tasks by either feature-based (e.g. ELMo) or fine-tuning (e.g. OpenAI GPT, BERT) methods.  We developed a new pre-training method based on BERT for pre-training on each visit of EHR so that the data with only one hospital visit can also be utilized. We revised BERT to fit EHR data in both input and pre-training objectives. To our best knowledge, \mname is the first model that leverages Transformers and language model pre-training techniques in healthcare domain. Compared with other supervised models, \mname can utilize discarded/unlabeled data more efficiently.
\item \textit{Medical ontology embedding with graph neural networks:}
We enhance the representation of medical codes via learning medical ontology embedding for each medical codes with graph neural networks. We then input the ontology embedding into a multi-layer Transformer~\cite{vaswani2017attention}  for BERT-style pre-training and fine-tuning.
\end{enumerate}

\section{Related Work}

\paragraph{Medication Recommendation.}
Medication Recommendation can be categorized into instance-based and longitudinal recommendation methods~\cite{shang2018gamenet}. Instance-based methods focus on current health conditions. Among them, Leap~\cite{zhang2017leap} formulates a multi-instance multi-label learning framework and proposes a variant of sequence-to-sequence model based on content-attention mechanism to predict combination of medicines given patient's diagnoses. Longitudinal-based methods leverage the temporal dependencies among clinical events, see ~\cite{choi2016retain,10.1371/journal.pone.0195024,lipton2015learning}. Among them, RETAIN~\cite{choi2016retain} uses a two-level neural attention model to detect influential past visits and significant clinical variables within those visits for improved medication recommendation.\\

\paragraph{Pre-training Techniques.}
The goal of pre-training techniques is to provide model training with good initializations. Pre-training has been shown extremely effective in various areas such as image classification \cite{hinton2006fast} and machine translation~\cite{ramachandran2016unsupervised}. The unsupervised pre-training can  be considered as a regularizer that supports better generalization from the training dataset~\cite{erhan2010does}. Recently,  language model pre-training techniques such as  \cite{peters2018deep,radford2018improving,devlin2018bert} have shown to largely improve the performance on multiple NLP tasks. As the most widely used one, BERT~\cite{devlin2018bert} builds on the Transformer~\cite{vaswani2017attention} architecture and improves the pre-training using a masked language model for bidirectional representation.
In this paper, we adapt the framework of BERT and pre-train our model on each visit of the EHR data to leverage the single-visit data that were not fit for training in other medication recommendation models. \\

\paragraph{Graph Neural Networks (GNN).}
GNNs are neural networks that learn node or graph representations from graph-structured data. Various graph neural networks have been proposed to encode the graph-structure information, including graph convolutional neural networks (GCN) \cite{kipf2016semi}, message passing networks (MPNN)~\cite{Gilmer2017}, graph attention networks (GAT)~\cite{velickovic2017graph}. GNNs have already been demonstrated useful on EHR modeling~\cite{choi2017gram,shang2018gamenet}. GRAM~\cite{choi2017gram} represented a medical concept as a combination of its ancestors in the medical ontology using an attention mechanism. It's different from \mname from two aspects as described in Section~\ref{sec:input}. Another work worth mentioning is GAMENet~\cite{shang2018gamenet}, which also used graph neural network to assist the medication recommendation task. However, GAMENet has a different motivation which results in using graph neural networks on drug-drug-interaction graphs instead of medical ontology.

\section{Problem Formalization}
\paragraph{Definition 1 (Longitudinal Patient Records).}
    In longitudinal EHR data, each patient can be represented as a sequence of multivariate observations: 
    $ \mathcal{X}^{(n)} = \{ \mathcal{X}_1^{(n)}, \mathcal{X}_2^{(n)}, \cdots, \mathcal{X}_{T^{(n)}}^{(n)} \} $ where 
    $n\in \{1,2,\ldots, N\}$, $N$ is the total number of patients; 
    $T^{(n)}$ is the number of visits of the $n^{th}$ patient. 
    Here we choose two main medical code to represent each visit $\mathcal{X}_t = \mathcal{C}_d^t \cup \mathcal{C}_m^t$ of a patient 
    which is a union set of corresponding diagnoses codes $\mathcal{C}_d^t \subset \mathcal{C}_d$ and medications codes $\mathcal{C}_m^t \subset \mathcal{C}_m$.
    For simplicity, we use $\mathcal{C}_\ast^t$ to indicate the unified definition for different type of medical codes and drop the superscript $(n)$ for a single patient whenever it is unambiguous.
    $\mathcal{C}_\ast$ denotes the medical code set and $|\mathcal{C}_\ast|$ the size of the code set.
    $c_\ast \in \mathcal{C}_\ast$ is the medical code. 

\paragraph{Definition 2 (Medical Ontology).} 
    Medical codes are usually categorized according to a tree-structured classification system
    such as ICD-9 ontoloy for diagnosis and ATC ontology for medication. 
    We use $\mathcal{O}_d, \mathcal{O}_m$ to denote the ontology for diagnosis and medication.
    Similarly, we use $\mathcal{O}_\ast$ to indicate the unified definition for different type of medical codes.
    In detial, $\mathcal{O}_\ast =  \overline{\mathcal{C}_\ast} \cup \mathcal{C}_\ast$ 
    where $\overline{\mathcal{C}_\ast}$ denotes the codes excluding leaf codes. 
    For simplicity, we define two function $pa(\cdot), ch(\cdot)$ which accept target
    medical code and return ancestors' code set and direct child code set.

\paragraph{Problem Definition (Medication Recommendation).}
    Given diagnosis codes $\mathcal{C}_d^t$ of the visit at time $t$, patient history $\mathcal{X}_{1:t} = \{\mathcal{X}_1, \mathcal{X}_2, \cdots, \mathcal{X}_{t-1}\}$,
    we want to recommend multiple medications by generating multi-label output $\hat{\bm{y}}_t \in \{0,1\}^{|\mathcal{C}_m|}$.
    

\section{Method}\label{sec:method}
\begin{figure*}[htp!]
\centering
\includegraphics[width=0.9\linewidth]{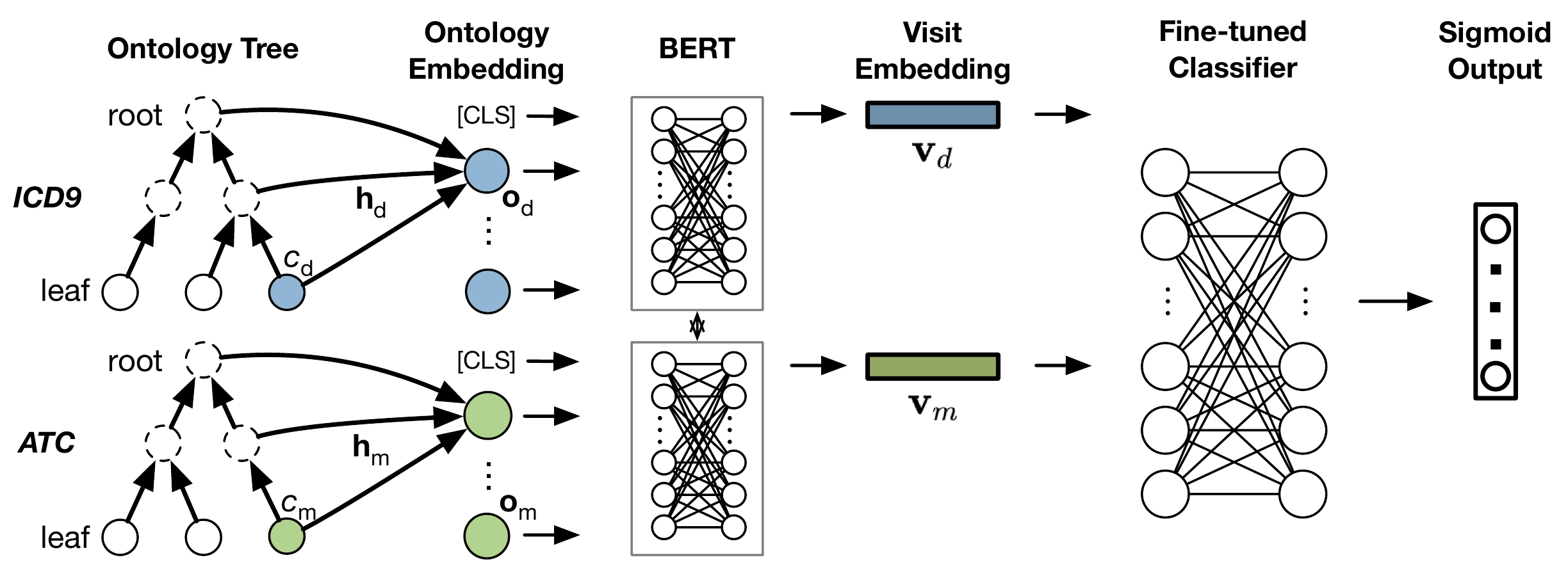}
\caption{The framework of \mname. It consists of three main parts: ontology embedding, BERT and fine-tuned classifier. Firstly, we derive ontology embedding for medical code laid in leaf nodes by cooperating ancestors information by Eq.~\ref{eq:ontoglogy_stage1} and \ref{eq:ontoglogy_stage2} based on graph attention networks (Eq.~\ref{eq:ontoglogy_gat},~\ref{eq:ontoglogy_gat_attn}). Then we input set of diagnosis and medication ontology embedding separately to shared weight BERT which is pre-trained using Eq.~\ref{eq:self_prediction},~\ref{eq:dual_prediction},~\ref{eq:pretraining_obj}. Finally, we concatenate the mean of all previous visit embeddings and the last visit embedding as input and fine-tune the prediction layers using Eq.~\ref{eq:fine_tuning_rx} for medication recommendation tasks.  }
\label{fig:framework}
\end{figure*}

The overall framework of \mname is described in Figure~\ref{fig:framework}. \mname first derives the initial embedding of medical codes from medical ontology using graph neural networks. Then, in order to fully utilize the rich EHR data, \mname constructs an adaptive BERT model on the discarded single-visit data for visit representation. Finally we add a prediction layer and fine-tune the model in the medication recommendation task. In the following we will describe \mname in detail. But firstly, we give a brief background of BERT especially for the two pre-training objectives which will be later adapted to EHR data in Section~\ref{sec:pre-training}.\\

\begin{table}[tb]
\centering
\resizebox{\columnwidth}{!}{
\begin{tabular}{l|l}
\hline
\bf Notation      & \bf Description                                            \\ \hline
$\mathcal{X}^{(n)}$ & longitudinal observations for n-th patient \\  
$\mathcal{C}_d, \mathcal{C}_m $    & diagnoses and medications codes set \\ 
$\mathcal{O}_d, \mathcal{O}_m $ & diagnoses and medications codes ontology \\ 
$\overline{\mathcal{C}_\ast} \subset \mathcal{O}_\ast$ & non-leaf medical codes of type $\ast$ \\\hline
$c_\ast \in \mathcal{O}_\ast$ & single medical code of type $\ast$ \\
$pa(c_\ast)$ & function retrieve $c_\ast$' ancestors' code  \\
$ch(c_\ast)$ & function retrieve $c_\ast$' direct children's code \\ \hline
$\bm{W}_e \in \mathbb{R}^{|\mathcal{O}_\ast| \times d}$ &  initial medical embedding matrix in $\mathcal{O}_\ast$ \\
$\bm{H}_e \in \mathbb{R}^{|\mathcal{O}_\ast| \times d}$ &  enhanced medical embeddings in \textbf{stage 1} \\
$\bm{o}_{c_\ast} \in \mathbb{R}^{d}$ & ontology embedding in \textbf{stage 2}\\ \hline
$\alpha_{i,j}^k $ & $k$-th attention between nodes $i,j$ \\
$\bm{W}^k \in \mathbb{R}^{d \times d}$ & $k$-th weight matrix applied to each node\\
$g(\cdot,\cdot,\cdot)$ & graph aggregator function \\ \hline
$\bm{v}_\ast^t$ & $t$-th visit embedding of type $\ast$ \\ 
$\hat{\bm{y}}_t \in \{0,1\}^{|\mathcal{C}_\ast|}$ & multi-label prediction \\ \hline
\end{tabular}}
\caption{Notations used in \mname.}\label{tab:notation}
\end{table}

\subsection{Background of BERT} 
Based on a multi-layer Transformer encoder~\cite{vaswani2017attention} (The transformer architecture has been ubiquitously used in many sequence modeling tasks recently, so we will not introduce the details here), BERT is pre-trained using two unsupervised tasks:
\begin{itemize}
    \item Masked Language Model. Instead of predicting words based on previous words, BERT randomly selects words to mask out and then predicts the original vocabulary IDs of the masked words from their (bidirectional) context.
    \item Next Sentence Prediction. Many of BERT's downstream tasks are predicting the relationships of two sentences, thus in the pre-training phase, BERT has am a binary sentence prediction task to predict whether one sentence is the next sentence of the other.
\end{itemize}
A typical input to BERT is as follows (~\cite{devlin2018bert}):
\begin{quote}
\textbf{Input} = [CLS] the man went to [MASK] store [SEP] he bought a gallon [MASK] milk  [SEP]\\
\textbf{Label} = IsNext
\end{quote}
where [CLS] is the first token of each sentence pair to represent the special classification embedding, i.e. the final state of this token is used as the aggregated sequence representation for classification tasks; [SEP] is used to separate two sentences; [MASK] is used to mask out the predicted words in the masked language model. Using this form, these inputs facilitate the two tasks described above, and they will also be used in our method description in the following section.

\subsection{Input Representation}
\label{sec:input}
The \mname model takes medical codes' ontology embeddings as input, and obtains intermediate representations from a Transformer encoder as the visit embeddings. It is then pre-trained on EHR from patients who only have one hospital visit. The derived encoder and visit embedding will be fed into a classifier and fine-tuned to make predictions.\\

\subsubsection{Ontology Embedding}
We constructed ontology embedding from diagnosis ontology $\mathcal{O}_d$ and medication ontology $\mathcal{O}_m$. 
Since the medical codes in raw EHR data can be considered as leaf nodes in these ontology trees, we can enhance the medical code embedding using graph neural networks (GNNs) to integrate the ancestors' information of these codes. Here we perform a two-stage procedure with a specially designed GNN for ontology embedding. \\

To start, we assign an initial embedding vector to every medical code $c_\ast \in \mathcal{O}_\ast$ with a learnable embedding matrix $\bm{W}_e \in \mathbb{R}^{|\mathcal{O}_\ast| \times d}$ where $d$ is the embedding dimension.\\

\paragraph{Stage 1.} For each non-leaf node $c_\ast \in \overline{\mathcal{C}_\ast}$, we obtain its enhanced medical embedding $\bm{h}_{c_\ast} \in \mathbb{R}^{d}$ as follows:
\begin{equation}\label{eq:ontoglogy_stage1}
    \bm{h}_{c_\ast} = g(c_\ast, ch(c_\ast), \bm{W}_e)
\end{equation}
where $g(\cdot,\cdot,\cdot)$ is an aggregation function which accepts the target medical code $c_\ast$, its direct child codes $ch(c_\ast)$ and initial embedding matrix. Intuitively, the aggregation function can pass and fuse information in target node from its direct children which result in the more related embedding of ancestor' code to child codes' embedding.\\

\paragraph{Stage 2.} After obtaining enhanced embeddings, we pass the enhance embedding matrix $\bm{H}_e \in \mathbb{R}^{|\mathcal{O}_\ast| \times d}$ back to get ontology embedding for leaf codes $c_\ast \in \mathcal{C}_\ast$ as follows:
\begin{equation}\label{eq:ontoglogy_stage2}
    \bm{o}_{c_\ast} = g(c_\ast, pa(c_\ast), \bm{H}_e)
\end{equation}
where $g(\cdot,\cdot,\cdot)$ accepts ancestor codes of target medical code $c_\ast$. Here, we use $pa(c_\ast)$ instead of $ch(c_\ast)$, since  utilizing the ancestors' embedding can indirectly associate all medical codes instead of taking each leaf code as independent input.

 The option for the aggregation function $g(\cdot,\cdot,\cdot)$ is flexible, including \textit{sum}, \textit{mean}. Here we choose the one from graph attention networks (GAT)~\cite{velickovic2017graph}, which has shown efficient embedding learning ability on graph-structured tasks, e.g., node classification and link prediction. In particular, we implement the aggregation function $g(\cdot,\cdot,\cdot)$ as follows (taking stage 2 for an example): 
\begin{equation}\label{eq:ontoglogy_gat}
    g (c_\ast, pa(c_\ast), \bm{H}_e) = \concat_{k=1}^K \sigma \left( \sum_{j \in \mathcal{N}_{c_\ast}} \alpha_{c_\ast,j}^k \bm{W}^k \bm{h}_j \right)
\end{equation}
where $\mathcal{N}_\ast$ is the neighboring nodes of $c_\ast$ including $c_\ast$ itself which depends on what stage we are on. Since we are on the stage 2 now, $\mathcal{N}_\ast=\{\{c_\ast\} \cup pa(c_\ast)\}$ (likewise for stage 1, $\mathcal{N}_\ast=\{\{c_\ast\} \cup ch(c_\ast)\}$), $\concat$ represents concatenation which enables the multi-head attention mechanism, $\sigma$ is a nonlinear activation function, $\bm{W}^k \in \mathbb{R}^{m \times d}$ is the weight matrix for input transformation, and $\alpha_{c_\ast,j}^k$ are the corresponding $k$-th normalized attention coefficients computed as follows:

\begin{equation}\label{eq:ontoglogy_gat_attn}
    \alpha_{c_\ast,j}^k = \frac{\exp{\left(\text{LeakyReLU}(\bm{a}^\intercal[\bm{W}^k \bm{h}_{c_\ast} || \bm{W}^k \bm{h}_j])\right)}}{\sum_{k \in \mathcal{N}_{c_\ast}} \exp{\left(\text{LeakyReLU}(\bm{a}^\intercal[\bm{W}^k \bm{h}_{c_\ast} || \bm{W}^k \bm{h}_k])\right)}}
\end{equation}
where $\bm{a} \in \mathbb{R}^{2m}$ is a learnable weight vector and LeakyReLU is a nonlinear function. (we assume $m = d/K$).\\

As shown in Figure~\ref{fig:framework}, we construct ICD-9 tree for diagnosis and ATC tree for medication using the same structure. Here the direction of arrow shows the information flow where ancestor nodes can get information from their direct children (in \textbf{stage 1}) and similarly leaf nodes can get information from their connected ancestors (in \textbf{stage 2}).\\
        
It is worth mentioning that our graph embedding method on medical ontology is different from GRAM~\cite{choi2017gram} from the following two aspects:
\begin{enumerate}
    \item \textit{Initialization}: we initialize all the node embeddings from a learnable embedding matrix, while GRAM learns them using Glove from the co-occurrence information.
    \item \textit{Updating}: we develop a two-step updating function for both leaf nodes and ancestor nodes; while in GRAM, only the leaf nodes are updated (as a combination of their ancestor nodes and themselves).
\end{enumerate}
        
\subsubsection{Visit Embedding}
Similar to BERT, we use a multi-layer Transformer architecture~\cite{vaswani2017attention} as our visit encoder. The model takes the ontology embedding as input and derive visit embedding $\bm{v}_\ast^t \in \mathbb{R}^d$ for a patient at $t$-th visit:
\begin{equation}\label{eq:visit_emb}
    \bm{v}_\ast^t = \mathrm{Transformer}(\{\textit{[CLS]}\} \cup \{\bm{o}_{c_\ast}^t| c_\ast \in \mathcal{C}_\ast^t\})[0]
\end{equation}
where \textit{[CLS]} is a special token as in BERT. It is put in the first position of each visit of type $\ast$ and its final state can be used as the representation of the visit. Intuitively, it is more reasonable to use Transformers as encoders (multi-head attention based architecture) than RNN or mean/sum to aggregate multiple medical embedding for visit embedding since the set of medical codes within one visit is not ordered. Note that symbol [SEP] is also ignored considering there is no clear separate among codes within one visit.\\

It is worth noting that our Transformer encoder is different from the original one in the position embedding part. Position embedding, as an important component in Transformers and BERT, is used to encode the position and order information of each token in a sequence. However, one big difference between language sentences and EHR sequences is that the medical codes within the same visit do not generally have an order, so we remove the position embedding in our model.

\subsection{Pre-training}\label{sec:pre-training}

\begin{figure}[t]
\centering
\includegraphics[width=\linewidth]{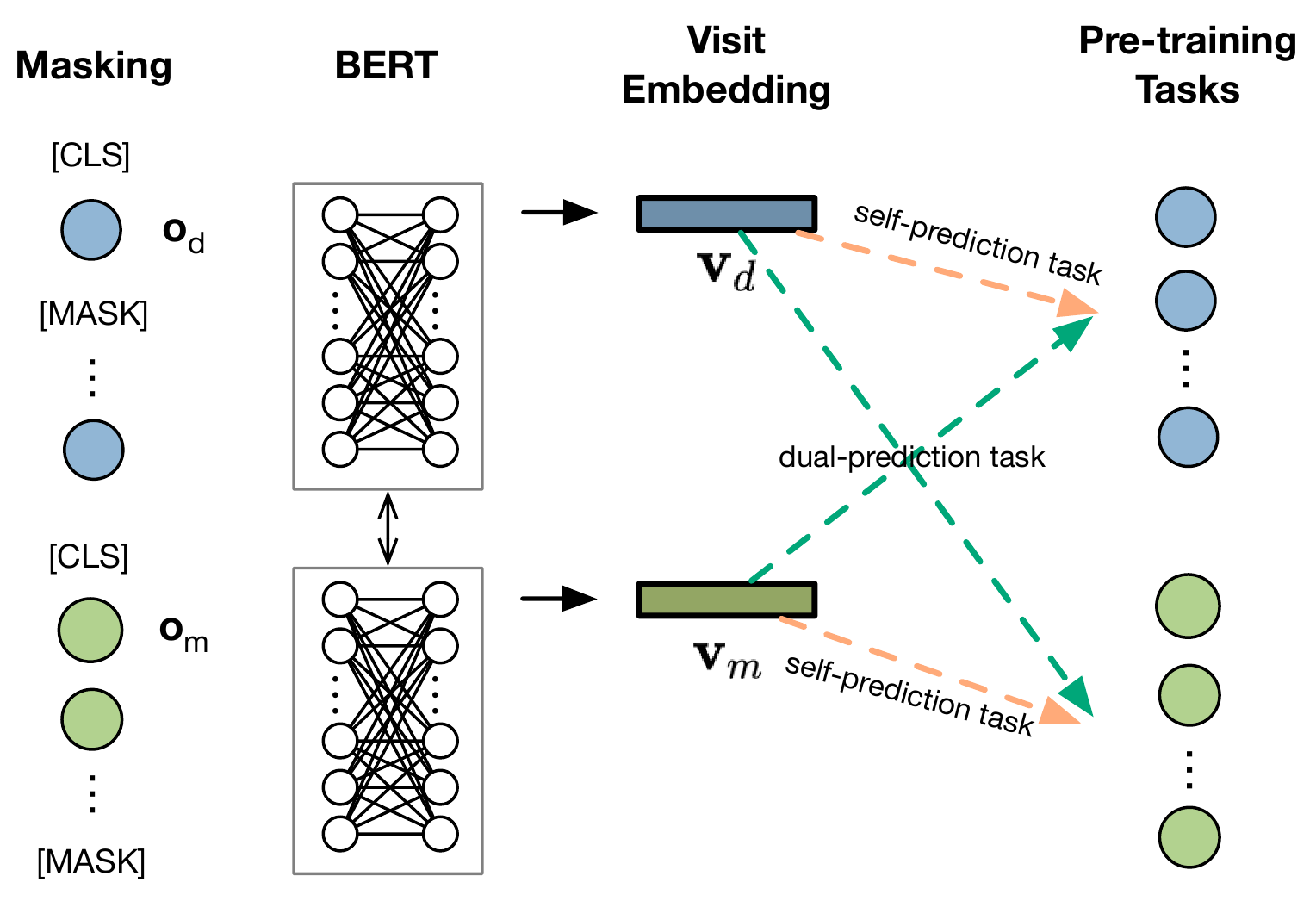}
\caption{Graphical illustration of pre-training procedure. We firstly randomly mask the input medical codes using a \textit{[MASK]} symbol. \textcolor{orange}{\textbf{Orange arrow}}: self-prediction task takes $\bm{v}_m$ or $\bm{v}_d$ as input to restore the original medical codes with the same type. \textcolor{green}{\textbf{Green arrow}}: dual-prediction task takes one type of visit embedding such as $\bm{v}_m$ or $\bm{v}_d$ and tries to predict the other type of medical codes.}
\label{fig:pre-training}
\end{figure}

We adapted the original BERT model to be more suitable for our data and task. In particular, we pre-train the model on each \textbf{single} EHR visit (within both single-visit EHR sequences and multi-visit EHR sequences). We modified the input and pre-training objectives of the BERT model: (1) For the input, we built the Transformer encoder on the GNN outputs, i.e. ontology embeddings, for visit embedding. For the original EHR sequence, it means essentially we combine the GNN model with a Transformer to become a new integrated encoder. In addition, we removed the position embedding as we explained before. (2) As for the pre-training procedures, we modified the original pre-training tasks i.e., Masked LM (language model) task and Next Sentence prediction task to \textbf{self-prediction task} and  \textbf{dual-prediction task}. The idea to conduct these tasks is to make the visit embedding absorb enough information about \textit{what it is made of} and \textit{what it is able to predict}. (note that we omit superscript $t$ as only single visit is used for pre-training procedure).\\

Thus, for the \textbf{self-prediction task}, we want the visit embedding $v_\ast$ to recover \textit{what it is made of}, i.e., the input medical codes $\mathcal{C}_\ast$ limited by the same type for each visit as follows:
\begin{equation}\label{eq:self_prediction}
\begin{aligned}
    &\mathcal{L}_{se}(\bm{v}_\ast, \mathcal{C}_\ast^{(n)}) = -\log p(\mathcal{C}_\ast^{(n)} | \bm{v}_\ast) \\
    & = - \sum_{c_\ast \in \mathcal{C}_\ast^{(n)}}\log p(c_\ast | \bm{v}_\ast) + \sum_{c_\ast \in \{\mathcal{C}_\ast \setminus \mathcal{C}_\ast^{(n)}\}}\log p(c_\ast | \bm{v}_\ast)
\end{aligned}
\end{equation}
where $\mathcal{C}_\ast^{(n)}$ is the medical codes set of $n$-th patient, $* \in \{d,m\}$ and we minimize the binary cross entropy loss $\mathcal{L}_{se}$. For instance, assume that the $n$-th patient takes 10 different medications out of total 100 medications which means $|\mathcal{C}_m^{(n)}|=10$ and $|\mathcal{C}_m|=100$. In such case, we instantiate and minimize $L_{se}(\bm{v}_m, \mathcal{C}_m^{(n)})$ to produce high probabilities among 10 taken medications captured by $- \sum_{c_\ast \in \mathcal{C}_m^{(n)}}\log p(c_\ast | \bm{v}_m)$ and lower the probabilities among 90 non-taken ones captured by $\sum_{c_\ast \in \{\mathcal{C}_m \setminus \mathcal{C}_m^{(n)}\}}\log p(c_\ast | \bm{v}_m)$. In practise, $\text{Sigmoid}(f(\bm{v}_\ast))$ should be transformed by applying a fully connected neural network $f(\cdot)$ with one hidden layer. With an analogy to the Masked LM task in BERT, we also used specific symbol [MASK] to randomly replace the original medical code $c_\ast \in \mathcal{C}_\ast$. So there are $15\%$ codes in $\mathcal{C}_\ast$ which will be replaced randomly and the model should have the ability to predict the masked code based on others.\\ 

Likewise, for the \textbf{dual-prediction task}, since the visit embedding $\bm{v}_\ast$ carries the information of medical codes of type $\ast$, we can further expect it has the ability to do more task-specific prediction as follows:
\begin{equation}\label{eq:dual_prediction}
\begin{aligned}
\mathcal{L}_{du} = -\log p(\mathcal{C}_d | \bm{v}_m) - \log p(\mathcal{C}_m | \bm{v}_d)
\end{aligned}
\end{equation}
where we use the same transformation function $\text{Sigmoid}(f_1(\bm{v}_m))$, $\text{Sigmoid}(f_2(\bm{v}_d))$\footnote{$f_1,f_2$ are the multiple layer perceptron (MLP) with one hidden layer} with different weight matrix to transform the visit embedding and optimize the binary cross entropy loss $\mathcal{L}_{du}$ expanded same as $\mathcal{L}_{se}$ in Eq.~\ref{eq:self_prediction}. This is a direct adaptation of the next sentence prediction task. In BERT, the next sentence prediction task facilitates the prediction of sentence relations, which is a common task in NLP. However, in healthcare, most predictive tasks do not have a sequence pair to classify. Instead, we are often interested in predicting unknown disease or medication codes of the sequence. For example, in medication recommendation, we want to predict multiple medications given only the diagnosis codes. Inversely, we can also predict unknown diagnosis given the medication codes.\\

Thus, our final pre-training optimization objective can simply be the combination of the aforementioned losses, as shown in Eq.~\ref{eq:pretraining_obj}. 
\begin{equation}\label{eq:pretraining_obj}
    \mathcal{L}_{pr} =  
    \mathcal{L}_{se}(\bm{v}_d, \mathcal{C}_d) + \mathcal{L}_{se}(\bm{v}_m, \mathcal{C}_m)
     + \mathcal{L}_{du}
\end{equation}
In practise, we could integrate using mini-batch technique to train on EHR data from all patients with a single visit.
\subsection{Fine-tuning}

\begin{figure}[t]
\centering
\includegraphics[width=\linewidth]{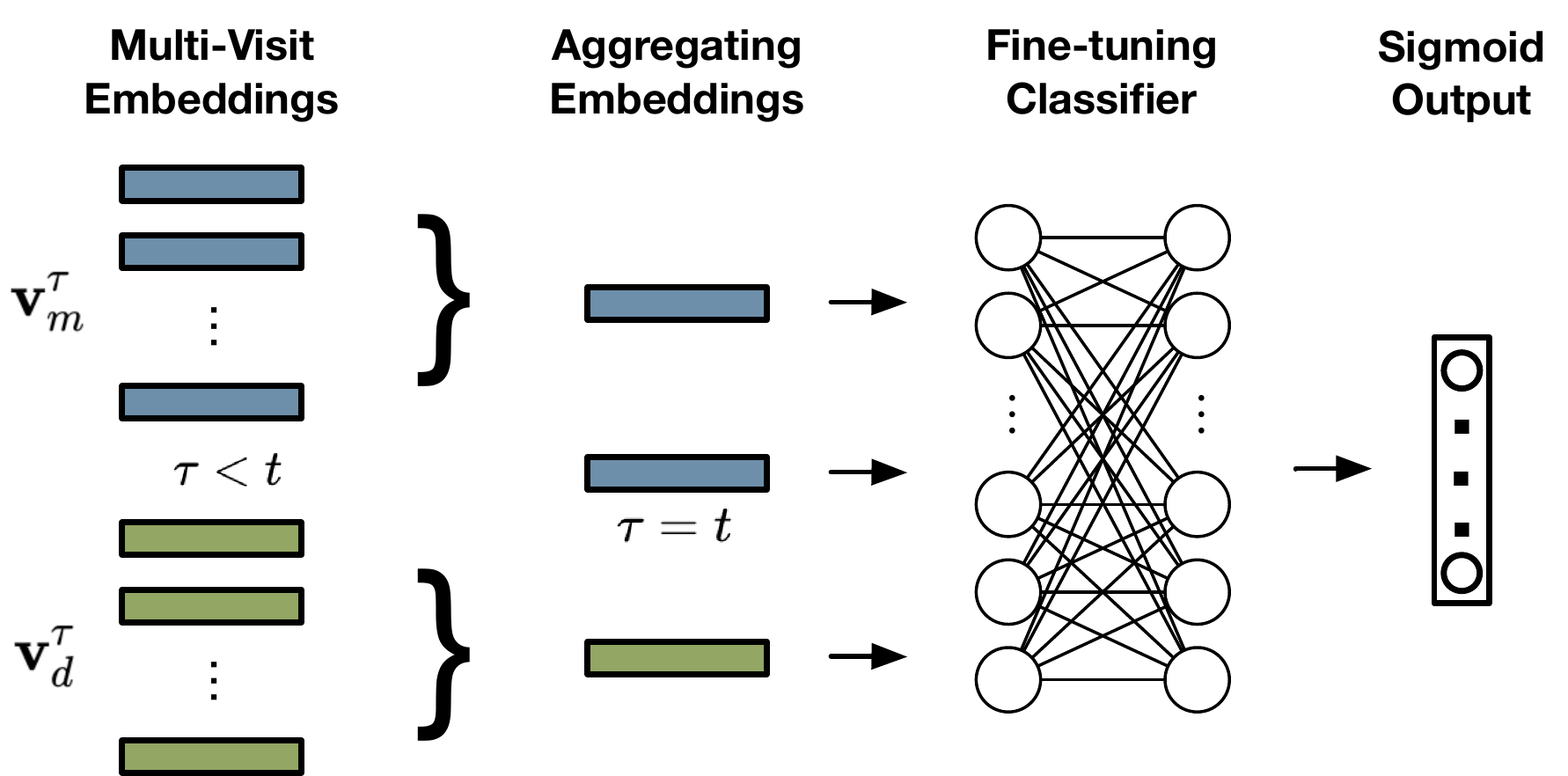}
\caption{Graphical illustration of fine-tuning procedure. We separately aggregate the visit embeddings produced by diagnoses and medications before $t$-visit. Then, these two aggregated visit embeddings are concatenated with diagnosis visit embedding at $t$-visit as input for fine-tuning.}
\label{fig:fine-tuning}
\end{figure}

After obtaining pre-trained visit representation for each visit,  for a prediction task on a multi-visit sequence data, we aggregate all the visit embedding and add a prediction layer for the medication recommendation task as shown in Figure.~\ref{fig:fine-tuning}. 
To be specific, from pre-training on all visits, we have a pre-trained Transformer encoder, which can then be used to get the visit embedding $\bm{v}_*^\tau$ at time $\tau$. The known diagnosis codes $\mathcal{C}_d^t$ at the prediction time $t$ is also represented using the same model as $\bm{v}_*^t$. Concatenating the mean of previous diagnoses visit embeddings and medication visit embeddings, also the last diagnoses visit embedding, we built an MLP based prediction layer to predict the recommended medication codes as in Equation~\ref{eq:pred1}.
\begin{equation}
   \bm{y}_t = \mathrm{Sigmoid}(\bm{W}_1 [(\frac{1}{t-1}\sum_{\tau < t} \bm{v}_d^\tau) || (\frac{1}{t-1}\sum_{\tau < t} \bm{v}_m^\tau) ||\bm{v}_d^t] + b) 
   \label{eq:pred1}
\end{equation}
where $\bm{W}_1 \in \mathbb{R}^{|\mathcal{C}_m| \times 3d} $ is a learnable transformation matrix.\\

Given the true labels $\hat{\bm{y}}_t$ at each time stamp $t$, the loss function for the whole EHR sequence (i.e. a patient) is
\begin{equation}\label{eq:fine_tuning_rx}
    \mathcal{L} = - \frac{1}{T-1} \sum_{t=2}^T (\bm{y}_t^\intercal \log(\hat{\bm{y}}_t) + (1 - \bm{y}_t^\intercal) \log(1 - \hat{\bm{y}}_t))
\end{equation}



\section{Experiment}
\subsection{Experimental Setting}
\subsubsection{Data}
We used EHR data from MIMIC-III~\cite{johnson2016mimic} and conducted all our experiments on a cohort where patients have more than one visit. We utilize data from patients with both single visit and multiple visits in the training dataset as pre-training data source (multi-visit data are split into visit slices and duplicate codes within a single visit are removed in order to avoid leakage of information). In this work, we transform the drug coding from NDC to ATC Third Level for using the ontology information. The statistics of the datasets are summarized in Table~\ref{tab:data}.\\
\vspace{-3mm}
\begin{table}[t]
\centering
\begin{tabular}{l|l|l}
\hline
Stats            & Single-Visit & Multi-Visit \\ \hline
\# of patients   & 30,745       & 6,350       \\
avg \# of visits & 1.00         & 2.36        \\ \hline
avg \# of dx     & 39           & 10.51       \\
avg \# of rx     & 52           & 8.80        \\ \hline
\# of unique dx  & 1,997        & 1,958       \\
\# of unique rx  & 323          & 145         \\ \hline
\end{tabular}
\caption{Statistics of the Data (dx for diagnosis, rx for medication).}\label{tab:data}
\end{table}

\subsubsection{Baselines}
We compared \mname\footnote{https://github.com/jshang123/G-Bert} with the following baselines. All methods are implemented in PyTorch~\cite{pytorch} and trained on an Ubuntu 16.04 with 8GB memory and Nvidia 1080 GPU.
\begin{enumerate}
\item \textbf{Logistic Regression (LR)} is logistic regression with L1/L2 regularization. Here we represent sequential multiple medical codes by sum of multi-hot vector of each visit. Binary relevance technique ~\cite{luaces2012binary} is used to handle multi-label output. 
\item \textbf{LEAP} \cite{zhang2017leap} is an instance-based medication combination recommendation method which formalizes the task in multi-instance and multi-label learning framework. It utilizes a encoder-decoder based model with attention mechanism to build complex dependency among diseases and medications.
\item \textbf{RETAIN} \cite{choi2016retain} makes sequential prediction of medication combination and diseases prediction based on a two-level neural attention model that detects influential past visits and clinical variables within those visits.
\item\textbf{GRAM}~\cite{choi2017gram} injects domain knowledge (ICD9 Dx code tree) to \textbf{tanh} via attention mechanism.
\item \textbf{GAMENet} \cite{shang2018gamenet} is the method to recommend accuracy and safe medication based on memory neural networks and graph convolutional networks by leveraging EHR data and Drug-Drug Interaction (DDI) data source. For fair comparison, we use a variant of GAMENet without DDI knowledge and procedure codes as input renamed as $\text{GAMENet}^-$.
\item \textbf{\mname} is our proposed model which integrated the GNN representation into Transformer-based visit encoder with pre-training on single-visit EHR data.
\end{enumerate}

We also evaluated 3 \mname variants for model ablation.
\begin{enumerate}
    \item \textbf{\mnamengnp}: We directly use medical embedding without ontology information as input and initialize the model's parameters without pre-training.
    \item \textbf{\mnameng}: We directly use medical embedding without ontology information as input with pre-training.
    \item \textbf{\mnamenp}: We use ontology information to get ontology embedding as input and initialize the model's parameters  without pre-training.
\end{enumerate}

\subsubsection{Metrics} To measure the prediction accuracy, we used Jaccard Similarity Score (Jaccard), Average F1 (F1) and Precision Recall AUC (PR-AUC).
Jaccard is defined as the size of the intersection divided by the size of the union of ground truth set $Y_t^{(k)}$ and predicted set $\hat{Y}_t^{(k)}$.
\begin{equation*}
\text{Jaccard} = \frac{1}{\sum_k^N \sum_t^{T_k} 1}\sum_k^N \sum_t^{T_k} \frac{|Y_t^{(k)} \cap \hat{Y}_t^{(k)}|}{|Y_t^{(k)} \cup \hat{Y}_t^{(k)}|}
\end{equation*}
where $N$ is the number of patients in test set and $T_k$ is the number of visits of the $k^{th}$ patient. \\

\subsubsection{Implementation Details}
We randomly divide the dataset into training, validation and testing set in a $0.6 : 0.2 : 0.2$ ratio. For \mname, the hyperparameters are adjusted on evaluation set: (1) GAT part: input embedding dimension as 75, number of attention heads as 4; (2) BERT part: hidden dimension as 300, dimension of position-wise feed-forward networks as 300, 2 hidden layers with 4 attention heads for each layer. Specially, we alternated the pre-training with 5 epochs and fine-tuning procedure with 5 epochs for 15 times to stabilize the training procedure.\\ 

For LR, we use the grid search over typical range of hyper-parameter to search the best hyperparameter values which result in L1 norm penalty with weight as $1.1$. For deep learning models, we implemented RNN using a gated recurrent unit (GRU)~\cite{cho2014properties} and utilize dropout with a probability of 0.4 on the output of embedding. We test several embedding choice for baseline methods and determine the dimension for medical embedding as 300 and thershold for final prediction as 0.3 for better performance. Training is done through Adam~\cite{DBLP:journals/corr/KingmaB14} at learning rate 5e-4. We fix the best model on evaluation set within 100 epochs and report the performance in test set.



\subsection{Results}

Table.~\ref{tab:exp_res} compares the performance on the medication recommendation task.
For variants of \mname, \mnamengnp performs worse compared with \mnameng and \mnamenp which demonstrate the effectiveness of using ontology information to get enhanced medical embedding as input and employ an unsupervised pre-training procedure on larger abundant data. Incorporating both hierarchical ontology information and pre-training procedure, the end-to-end model \mname has more capacity and achieve comparable results with others.

\begin{table}[ht]
\centering
\resizebox{\columnwidth}{!}{
\begin{tabular}{l|c|c|c|c}
\hline
 Methods &  Jaccard & PR-AUC  & F1 & \# of parameters \\ \hline
 LR &0.4075 & 0.6716 &0.5658 & - \\
 GRAM & 0.4176& 0.6638 & 0.5788& 3,763,668 \\
 LEAP & 0.3921& 0.5855& 0.5508& 1,488,148 \\
 RETAIN & 0.4456 & 0.6838 & 0.6064 & 2,054,869 \\
 GAMENet$^-$ & 0.4401& 0.6672& 0.5996& 5,518,646 \\
 GAMENet & 0.4555& 0.6854& 0.6126& 5,518,646 \\ \hline
\mnamengnp & 0.4186& 0.6649& 0.5796& 2,634,145 \\
\mnameng & 0.4299& 0.6771& 0.5903& 2,634,145 \\
\mnamenp & 0.4236& 0.6704& 0.5844& 3,034,045 \\
\mname & \bf 0.4565 & \bf 0.6960&  \bf 0.6152& 3,034,045 \\ \hline
\end{tabular}}
\caption{Performance on Medication Recommendation Task.}\label{tab:exp_res}
\end{table}

As for baseline models, LR and Leap are worse than our most basic model (\mnamengnp) in terms of most metrics. Comparing \mnamenp and GRAM, which both used medical ontology information without pre-training, the scores of our \mnamenp is slightly higher in all metrics. This can demonstrate the validness of using Transformer encoders and the specific prediction layer for medication recommendation. Our final model \mname is also better than the attention based model, RETAIN, and the recently published state-of-the-art model, GAMENet. Specifically, even adding the extra information of DDI knowledge and procedure codes, GAMENet still performs worse than \mname.\\

In addition, we visualized the pre-training medical code embeddings of \mnameng and \mname to show the effectiveness of ontology embedding using online embedding projector \footnote{https://projector.tensorflow.org/} shown in (\url{https://raw.githubusercontent.com/jshang123/G-Bert/master/saved/tsne.png/}).




\section{Conclusion}

In this paper we proposed a pre-training model named \mname for medical code representation and medication recommendation. To our best knowledge, \mname is the first that utilizes language model pre-training techniques in healthcare domain. It adapted BERT to the EHR data and integrated medical ontology information using graph neural networks. By additional pre-training on the EHR from patients who only have one hospital visit which are generally discarded before model training, \mname outperforms all baselines in prediction accuracy on medication recommendation task. One direction for the future work is to add more auxiliary and structural tasks to improve the ability of code representaion. Another direction may be to adapt our model to be suitable for even larger datasets with more heterogeneous modalities.

\section*{Acknowledgments}
This work was supported by the National Science Foundation award IIS-1418511, CCF-1533768 and IIS-1838042, the National Institute of Health award 1R01MD011682-01 and R56HL138415.


\begin{thebibliography}{}

\bibitem[\protect\citeauthoryear{Baytas \bgroup \em et al.\egroup
  }{2017}]{Baytas:2017:PSV:3097983.3097997}
Inci~M. Baytas, Cao Xiao, Xi~Zhang, Fei Wang, Anil~K. Jain, and Jiayu Zhou.
\newblock Patient subtyping via time-aware lstm networks.
\newblock In {\em Proceedings of the 23rd ACM SIGKDD International Conference
  on Knowledge Discovery and Data Mining}, 2017.

\bibitem[\protect\citeauthoryear{Cho \bgroup \em et al.\egroup
  }{2014}]{cho2014properties}
Kyunghyun Cho, Bart Van~Merri{\"e}nboer, Dzmitry Bahdanau, and Yoshua Bengio.
\newblock On the properties of neural machine translation: Encoder-decoder
  approaches.
\newblock {\em arXiv preprint arXiv:1409.1259}, 2014.

\bibitem[\protect\citeauthoryear{Choi \bgroup \em et al.\egroup
  }{2016}]{choi2016retain}
Edward Choi, Mohammad~Taha Bahadori, Jimeng Sun, Joshua Kulas, Andy Schuetz,
  and Walter Stewart.
\newblock Retain: An interpretable predictive model for healthcare using
  reverse time attention mechanism.
\newblock In {\em Advances in Neural Information Processing Systems}, pages
  3504--3512, 2016.

\bibitem[\protect\citeauthoryear{Choi \bgroup \em et al.\egroup
  }{2017}]{choi2017gram}
Edward Choi, Mohammad~Taha Bahadori, Le~Song, Walter~F Stewart, and Jimeng Sun.
\newblock Gram: Graph-based attention model for healthcare representation
  learning.
\newblock In {\em SIGKDD}, 2017.

\bibitem[\protect\citeauthoryear{Choi \bgroup \em et al.\egroup
  }{2018}]{NIPS2018_7706}
Edward Choi, Cao Xiao, Walter Stewart, and Jimeng Sun.
\newblock Mime: Multilevel medical embedding of electronic health records for
  predictive healthcare.
\newblock In S.~Bengio, H.~Wallach, H.~Larochelle, K.~Grauman, N.~Cesa-Bianchi,
  and R.~Garnett, editors, {\em Advances in Neural Information Processing
  Systems 31}, pages 4547--4557. Curran Associates, Inc., 2018.

\bibitem[\protect\citeauthoryear{Devlin \bgroup \em et al.\egroup
  }{2018}]{devlin2018bert}
Jacob Devlin, Ming-Wei Chang, Kenton Lee, and Kristina Toutanova.
\newblock Bert: Pre-training of deep bidirectional transformers for language
  understanding.
\newblock {\em arXiv preprint arXiv:1810.04805}, 2018.

\bibitem[\protect\citeauthoryear{Erhan \bgroup \em et al.\egroup
  }{2010}]{erhan2010does}
Dumitru Erhan, Yoshua Bengio, Aaron Courville, Pierre-Antoine Manzagol, Pascal
  Vincent, and Samy Bengio.
\newblock Why does unsupervised pre-training help deep learning?
\newblock {\em Journal of Machine Learning Research}, 11(Feb):625--660, 2010.

\bibitem[\protect\citeauthoryear{Gilmer \bgroup \em et al.\egroup
  }{2017}]{Gilmer2017}
J.~Gilmer, S.S. Schoenholz, P.F. Riley, O.~Vinyals, and G.E. Dahl.
\newblock Neural message passing for quantum chemistry.
\newblock In {\em ICML}, 2017.

\bibitem[\protect\citeauthoryear{Hinton \bgroup \em et al.\egroup
  }{2006}]{hinton2006fast}
Geoffrey~E Hinton, Simon Osindero, and Yee-Whye Teh.
\newblock A fast learning algorithm for deep belief nets.
\newblock {\em Neural computation}, 18(7):1527--1554, 2006.

\bibitem[\protect\citeauthoryear{Johnson \bgroup \em et al.\egroup
  }{2016}]{johnson2016mimic}
Alistair~EW Johnson, Tom~J Pollard, Lu~Shen, H~Lehman Li-wei, Mengling Feng,
  Mohammad Ghassemi, Benjamin Moody, Peter Szolovits, Leo~Anthony Celi, and
  Roger~G Mark.
\newblock Mimic-iii, a freely accessible critical care database.
\newblock {\em Scientific data}, 3:160035, 2016.

\bibitem[\protect\citeauthoryear{Kingma and
  Ba}{2014}]{DBLP:journals/corr/KingmaB14}
Diederik~P. Kingma and Jimmy Ba.
\newblock Adam: {A} method for stochastic optimization.
\newblock {\em CoRR}, abs/1412.6980, 2014.

\bibitem[\protect\citeauthoryear{Kipf and Welling}{2017}]{kipf2016semi}
Thomas~N Kipf and Max Welling.
\newblock Semi-supervised classification with graph convolutional networks.
\newblock In {\em ICLR}, 2017.

\bibitem[\protect\citeauthoryear{Lipton \bgroup \em et al.\egroup
  }{2015}]{lipton2015learning}
Zachary~C Lipton, David~C Kale, Charles Elkan, and Randall Wetzel.
\newblock Learning to diagnose with lstm recurrent neural networks.
\newblock {\em arXiv preprint arXiv:1511.03677}, 2015.

\bibitem[\protect\citeauthoryear{Luaces \bgroup \em et al.\egroup
  }{2012}]{luaces2012binary}
Oscar Luaces, Jorge D{\'\i}ez, Jos{\'e} Barranquero, Juan~Jos{\'e} del Coz, and
  Antonio Bahamonde.
\newblock Binary relevance efficacy for multilabel classification.
\newblock {\em Progress in Artificial Intelligence}, 1(4):303--313, 2012.

\bibitem[\protect\citeauthoryear{Ma \bgroup \em et al.\egroup
  }{2018}]{ma2018health}
Tengfei Ma, Cao Xiao, and Fei Wang.
\newblock Health-atm: A deep architecture for multifaceted patient health
  record representation and risk prediction.
\newblock In {\em Proceedings of the 2018 SIAM International Conference on Data
  Mining}, pages 261--269. SIAM, 2018.

\bibitem[\protect\citeauthoryear{Paszke \bgroup \em et al.\egroup
  }{2017}]{pytorch}
Adam Paszke, Sam Gross, Soumith Chintala, Gregory Chanan, Edward Yang, Zachary
  DeVito, Zeming Lin, Alban Desmaison, Luca Antiga, and Adam Lerer.
\newblock Automatic differentiation in pytorch.
\newblock 2017.

\bibitem[\protect\citeauthoryear{Peters \bgroup \em et al.\egroup
  }{2018}]{peters2018deep}
Matthew Peters, Mark Neumann, Mohit Iyyer, Matt Gardner, Christopher Clark,
  Kenton Lee, and Luke Zettlemoyer.
\newblock Deep contextualized word representations.
\newblock In {\em NAACL}, volume~1, pages 2227--2237, 2018.

\bibitem[\protect\citeauthoryear{Radford \bgroup \em et al.\egroup
  }{2018}]{radford2018improving}
Alec Radford, Karthik Narasimhan, Tim Salimans, and Ilya Sutskever.
\newblock Improving language understanding by generative pre-training.
\newblock 2018.

\bibitem[\protect\citeauthoryear{Ramachandran \bgroup \em et al.\egroup
  }{2016}]{ramachandran2016unsupervised}
Prajit Ramachandran, Peter~J Liu, and Quoc~V Le.
\newblock Unsupervised pretraining for sequence to sequence learning.
\newblock {\em arXiv preprint arXiv:1611.02683}, 2016.

\bibitem[\protect\citeauthoryear{Shang \bgroup \em et al.\egroup
  }{2019}]{shang2018gamenet}
Junyuan Shang, Cao Xiao, Tengfei Ma, Hongyan Li, and Jimeng Sun.
\newblock Gamenet: Graph augmented memory networks for recommending medication
  combination.
\newblock {\em AAAI}, 2019.

\bibitem[\protect\citeauthoryear{Vaswani \bgroup \em et al.\egroup
  }{2017}]{vaswani2017attention}
Ashish Vaswani, Noam Shazeer, Niki Parmar, Jakob Uszkoreit, Llion Jones,
  Aidan~N Gomez, {\L}ukasz Kaiser, and Illia Polosukhin.
\newblock Attention is all you need.
\newblock In {\em Advances in Neural Information Processing Systems}, pages
  5998--6008, 2017.

\bibitem[\protect\citeauthoryear{Velickovic \bgroup \em et al.\egroup
  }{2017}]{velickovic2017graph}
Petar Velickovic, Guillem Cucurull, Arantxa Casanova, Adriana Romero, Pietro
  Lio, and Yoshua Bengio.
\newblock Graph attention networks.
\newblock {\em arXiv preprint arXiv:1710.10903}, 1(2), 2017.

\bibitem[\protect\citeauthoryear{Xiao \bgroup \em et al.\egroup
  }{2018a}]{doi:10.1093/jamia/ocy068}
Cao Xiao, Edward Choi, and Jimeng Sun.
\newblock Opportunities and challenges in developing deep learning models using
  electronic health records data: a systematic review.
\newblock {\em Journal of the American Medical Informatics Association}, 2018.

\bibitem[\protect\citeauthoryear{Xiao \bgroup \em et al.\egroup
  }{2018b}]{10.1371/journal.pone.0195024}
Cao Xiao, Tengfei Ma, Adji~B. Dieng, David~M. Blei, and Fei Wang.
\newblock Readmission prediction via deep contextual embedding of clinical
  concepts.
\newblock {\em PLOS ONE}, 13(4):1--15, 04 2018.

\bibitem[\protect\citeauthoryear{Zhang \bgroup \em et al.\egroup
  }{2017}]{zhang2017leap}
Yutao Zhang, Robert Chen, Jie Tang, Walter~F Stewart, and Jimeng Sun.
\newblock Leap: Learning to prescribe effective and safe treatment combinations
  for multimorbidity.
\newblock In {\em SIGKDD}, pages 1315--1324, 2017.

\end{thebibliography}

\end{document}